\documentclass{bmvc2k}

\usepackage{times}
\usepackage{epsfig}
\usepackage{graphicx}
\usepackage{amsmath}
\usepackage{amssymb}

\usepackage{enumitem}
\usepackage{bbm}
\usepackage{tabularx}

\usepackage{cuted}
\usepackage{capt-of}

\title{LARNet: Latent Action Representation for Human Action Synthesis}

\addauthor{Naman Biyani}{namanb@iitk.ac.in}{1}
\addauthor{Aayush J Rana}{aayushjr@knights.ucf.edu}{2}
\addauthor{Shruti Vyas}{shruti@crcv.ucf.edu}{2}
\addauthor{Yogesh S Rawat}{yogesh@crcv.ucf.edu}{2}

\addinstitution{
 IIT Kanpur\\
 Kanpur, India
}
\addinstitution{
 Center for Research in Computer Vision\\
 University of Central Florida,\\
 Florida, USA
}

\runninghead{Biyani et al.}{LARNet: Latent Representation for Action Synthesis}

\begin{document}

\maketitle

%%%%%%%%% ABSTRACT
\begin{abstract}

We present LARNet, a novel end-to-end approach for generating human action videos. A joint generative modeling of appearance and dynamics to synthesize a video is very challenging and therefore recent works in video synthesis have proposed to decompose these two factors. However, these methods require a driving video to model the video dynamics. In this work, we propose a generative approach instead, which explicitly learns action dynamics in latent space avoiding the need of a driving video during inference. The generated action dynamics is integrated with the appearance using a recurrent hierarchical structure which induces motion at different scales to focus on both coarse as well as fine level action details. In addition, we propose a novel mix-adversarial loss function which aims at improving the temporal coherency of synthesized videos. We evaluate the proposed approach on four real-world human action datasets demonstrating the effectiveness of the proposed approach in generating human actions. Code available at \url{https://github.com/aayushjr/larnet}.

% say how we do it what is so unique about it...
% explicitly decomposing appearance and motion, propose novel generative modeling for learning latent action representation... the learned action representation is integrated with the appearance in a hierarchical approach to generate the action video...

% In addition, we propose a novel loss function [NAME IT] specifically designed to focus on motion in the generated video...

\iffalse

%%% NOTES
basically, we will have three main focus points
1 - the unique way in which we generate video, decompose appearance and motion, generative approach for latent action representation, which is novel, we need to show what if we don't use it and we have results for those...
2 - the hierarchical recurrent motion integrator, we will have show its effect and I think we have that in the ablations... what if we have fewer levels, what if we don't use it... we should have these 3-4 ablations...
3 - The last will be the loss function, I am assuming the frame different will be the one, if not the mix-adversarial loss... we will have to show its effect clearly...

\fi

\end{abstract}

\section{Introduction}

%%%
\iffalse
little motivation and existing works...
\quad
what are the limitations and challenges which we are addressing...

why decomposing appearance and motion is important and how it has been used so far... but it is implicit learning and there is no direct supervision... motion transfer kind of works show good results, but they require a driving video which is a big limitation... we address both of these issues...

In intro only focus on explicit disentanglement of appearance and motion... how this is different from existing works, what are the benefits, hierarchical recurrent integration of appearance and motion to learn both coarse as well as fine-grained features, and finally the novel loss function focused on temporal coherence... 

conclude with contributions...
\fi
%%%

\begin{figure*}
\begin{center}
% \fbox{\rule{0pt}{2in} \rule{.9\linewidth}{0pt}}
\includegraphics[width=0.75\linewidth]{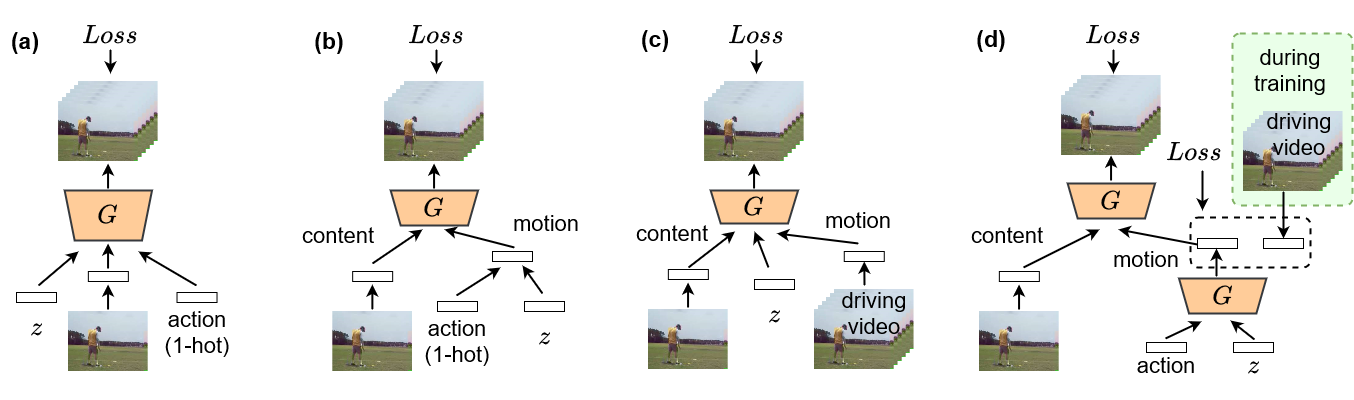}
\end{center}
  \caption{\small{Overview of different approaches for video synthesis. (a) joint generative modeling \cite{vondrick2016generating, saito2017temporal, saito2018tganv2, WANG_2020_WACV, kumar2019videoflow}, (b) implicit disentanglement of content and motion \cite{ohnishi2017hierarchical,villegas2017decomposing,Tulyakov_2018_CVPR_mocogan,wang2020g3an}, (c) motion transfer from a driving video \cite{ohnishi2017hierarchical,hu2018video,chan2019everybody}, and (d) proposed approach, explicit disentanglement of content and motion without requiring any driving video during inference.}}
\label{fig:teaser}
\end{figure*}

Video generation is a challenging problem with a lot of applications in robotics \cite{lee2018stochastic,finn2016unsupervised}, augmented reality \cite{wang2019few, wang2018video}, data augmentation \cite{siarohin2019animating,gur2020hierarchical,zhang2019self,bansal2018recycle}, and action imitation \cite{clark2019efficient,Tulyakov_2018_CVPR_mocogan,mathieu2015deep,vondrick2016generating,walker2017pose,liang2017dual,acharya2018towards}. 
%In recent years, there has been a great interest from the research community in this problem domain \cite{wang2020g3an,clark2019efficient,saito2018tganv2,Tulyakov_2018_CVPR_mocogan,saito2017temporal,vondrick2016generating}. 
It has different variations, such as video prediction \cite{kwon2019predicting, ye2019compositional, kratzwald2017improving}, video synthesis \cite{wang2019few,wang2018video}, video interpolation \cite{tran2020efficient, park2020vidode}, and super-resolution \cite{lucas2019generative,johnson2016perceptual,demir2021tinyvirat}. 
% Generating realistic videos is still a challenging problem due to spatio-temporal complexity and high computational requirements \cite{clark2019efficient,saito2018tganv2}. 
% The use of prior conditions, such as action class \cite{wang2020g3an,vondrick2016generating}, pose information \cite{walker2017pose, yan2018spatial}, previous frames \cite{kwon2019predicting, ye2019compositional}, etc., leads to a better quality when compared with the videos generated without any prior information in an unconditional setting \cite{vondrick2016generating,saito2017temporal,Tulyakov_2018_CVPR_mocogan}.
% In this work, we focus on generating human actions conditioned on image of an actor and a target action. Although a generative approach can be used to predict the future frames in a video, the problem of predicting future frames from a single input image is ill-posed. Given an input frame, a generated video can have a different sequence of future frames depending on the performed action. To address this, the existing approaches utilize a stochastic noise \cite{vondrick2016generating} or some kind of prior condition \cite{wang2020g3an,Tulyakov_2018_CVPR_mocogan} for synthesizing future frames. 
In this work, we focus on generating human actions conditioned on image of an actor and a target action. A generative approach can be used to predict the future frames in a video, but the problem of predicting future frames from a single input image is ill-posed. Given an input frame, a generated video can have a different sequence of future frames depending on the variation in the performed action. A use of stochastic noise can address this limitation \cite{vondrick2016generating, saito2017temporal, saito2018tganv2, WANG_2020_WACV, kumar2019videoflow}, but a joint generative modeling of appearance and motion for video synthesis is very challenging (Figure \ref{fig:teaser}(a)). 

%%%
% \vspace{-0.2cm}
%%%

To address this, the existing approaches utilize a decomposition of appearance and motion \cite{ohnishi2017hierarchical,villegas2017decomposing,Tulyakov_2018_CVPR_mocogan,wang2020g3an,li2021pose} (Figure \ref{fig:teaser}(b)). This enables the model to independently learn the variations in the performed action and helps in video synthesis. With a similar motivation, there are approaches explicitly using a prior motion which avoids the need of motion modelling and simplifying the complexity of video synthesis (Figure \ref{fig:teaser}(c)). This approach has been found very effective for video synthesis by motion transfer \cite{ohnishi2017hierarchical,hu2018video,chan2019everybody}.

We want to benefit from both these approaches (Figure \ref{fig:teaser} (b) and (c)) for conditional human action synthesis. However, each of these come with their own limitations. In the first approach, the decomposition of appearance and motion is not explicit as the only supervision comes from the generated video which limits the potential of this disentanglement. And in the second approach, the explicit use of motion information requires a synchronized driving video during inference which also restricts the motion generating capability of the model.

We propose LARNet, a generative framework which attempts to benefit from both these approaches and simultaneously overcome the above two limitations. LARNet explicitly models the action dynamics in latent space by approximating it to motion from real action videos. This enables effective decomposition of appearance and motion while avoiding the need of any driving video during inference (Figure \ref{fig:teaser}(d)). The disentangled appearance and motion features needs to be integrated effectively for video synthesis. LARNet utilizes a recurrent hierarchical structure for this integration focusing at different scales for capturing both coarse as well fine-level action details. 
% In addition, we propose XYZ-loss which helps in generating a video with coherent temporal action dynamics. 
The proposed method is trained end-to-end in an adversarial framework, optimizing multiple objectives. We make the following novel contributions in this work,
\begin{enumerate}[itemsep=0ex]
    \item We propose a generative approach for human action synthesis that leverages the decomposition of content and motion by explicit modeling of action dynamics.
    \item We propose a hierarchical recurrent motion integration approach which operates at multiple scales focusing on both coarse level and fine level details.
    \item We propose mix-adversarial loss, a novel objective function for video synthesis which aims at improving the temporal coherency in the synthesized videos.
\end{enumerate}

We validate our approach on several real-world human action datasets, showing its effectiveness in generating human action videos. % We demonstrate: (a) the ability of our proposed action representation learning framework for action synthesis; (b) the benefits of recurrent hierarchical motion integration; and (c) the capability of proposed XYZ-loss to generate videos with coherent motion. 

% In this work, we focus on the problem of video synthesis from conditioned on first frame and action annotations.
% The presence of novel views makes the video synthesis task more complex as both the action and appearance vary significantly with the change in
% viewpoint.
% There has been some work in action-conditioned video generation in which
% the focus is on improving video generation using classes, also in action-conditioned video synthesis, no prior uses action annotations properly although annotations are available for most video datasets.

% We propose an end-to-end deep adversarial framework to solve the problem of action-conditioned video synthesis. The proposed framework takes a the first frame of target video as an appearance prior and action annotation as a action prior and the target video is synthesized using . We use a \ recurrent transformer network, which takes the latent action features generated from annotations in an unsupervised approach and recurrently transforms the appearance to generate a sequence of target action features in the latent space.  Moreover, we propose a hierarchical structure, which enables the network to perform the transformation at different feature scales. We use a novel dual discriminator approach- a video discriminator and a action discriminator. We also use a contrastive loss in action representation learning.

% introduce the problem...

% what is the challenge...

% existing works...

% how we solve this...

% contributions...

\section{Related Work}

\paragraph{Video Prediction}

Video prediction task predicts future frames by conditioning on the input frame(s)\cite{ye2019compositional,kratzwald2017improving, kumar2019videoflow, franceschi2020stochastic, xu2020video}.
%\YSR{cite all the video prediction papers from the reviews...} . 
Using future frames as ground-truth leads to conditioned supervised learning approach which gives better results in contrast to unconditional video generation \cite{kratzwald2017improving,mathieu2015deep,finn2016unsupervised,chao2017forecasting}. GAN based approaches often relies on a sequence of input frames as priors to reduce ambiguity \cite{ye2019compositional,fragkiadaki2017motion,denton2018stochastic,walker2017pose}. Our approach uses only the first input frame and action class name as prior for the prediction task similar to \cite{vondrick2016generating,kratzwald2017improving}.

% \vspace{-0.2cm}

%talk about future frame and video prediction, when a sequence of frames are provided as input... two categories frame prediction, and video prediction... you can say that we already have past motion information and we need to predict how it evolves over time... another variation is frame interpolation, which is different where we need to determine intermediate frames to predict a smooth motion... our focus is different where we do not have a sequence of frames a input and therefore no prior motion information... we just use action class name to predict motion...

\paragraph{Video Synthesis}

Although GANs have been successful in image synthesis task \cite{brock2018large,wang2017high,choi2018stargan,zhang2018self,ledig2017photo}, synthesizing a high resolution realistic video is still challenging due to the temporal complexity and resource requirements \cite{clark2019efficient, Tulyakov_2018_CVPR_mocogan, vondrick2016generating, saito2018tganv2,vyas2020multi}. GANs use RNN architectures \cite{li2019storygan,Tulyakov_2018_CVPR_mocogan}, progressive generative models \cite{duan2019cascade,li2018video,demir2021tinyvirat} or decoupled two-stream approach \cite{sun2018two,vondrick2016generating} to address this. Unconditional video GANs rely on various forms to improve on spatio-temporal consistency such as random noise input  \cite{Tulyakov_2018_CVPR_mocogan,clark2019efficient}, two-stream learning \cite{Tulyakov_2018_CVPR_mocogan,vondrick2016generating}, multi-scale approach \cite{saito2018tganv2} and increasing computing power \cite{clark2019efficient}.
%The unconditional video GANs attempt to generate videos using only random noise as the input, resulting in a challenging training scenario \cite{Tulyakov_2018_CVPR_mocogan,clark2019efficient}. \cite{Tulyakov_2018_CVPR_mocogan,vondrick2016generating} apply a two stream approach to learn different components of the video. \cite{saito2018tganv2} adapts a multi-scale approach by stacking multiple sub-generators to generate complex features. \cite{clark2019efficient} adapts \cite{brock2018large} for large scale video generation based on complex datasets using massive computing powers.
In contrast, conditional video synthesis task is able to generate higher quality videos and easily learn the data distribution. Conditional GANs have many variants that use text \cite{li2019storygan,balaji2019conditional}, speech \cite{Chen_2018_ECCV,zhou2019talking,mittal2020animating}, class label \cite{Tulyakov_2018_CVPR_mocogan, clark2019efficient, wang2020g3an}, pose information \cite{walker2017pose,yang2018pose,li2021pose}, semantics \cite{wang2018video,wang2019few}, the entire video \cite{bansal2018recycle,vyas2020multi,schatz2020recurrent,shiraz2021novel} or only first frame \cite{vondrick2016generating,kratzwald2017towards}. 
%These methods have shown that using conditional input improves the generation quality compared to their unconditional counterparts.
Our approach uses first frame and class label embedding for conditioning and is evaluated against prior approaches \cite{vondrick2016generating,Tulyakov_2018_CVPR_mocogan,wang2020g3an, siarohin2019animating_monkeynet, WANG_2020_WACV}. 

%\cite{wang2018video,wang2019few} use videos with semantic frames to produce realistic images in it. \cite{vondrick2016generating,kratzwald2017towards} use first frame as condition while \cite{Tulyakov_2018_CVPR_mocogan,wang2020g3an} use class label as condition to generate videos. \cite{bansal2018recycle} uses entire video as condition and translates a video from one domain to another. Our approach is closely related to \cite{vondrick2016generating,Tulyakov_2018_CVPR_mocogan,wang2020g3an} as it uses first input frame and class label embedding as conditional input and we perform direct comparisons with these methods. 
% \vspace{-0.2cm}

\paragraph{Motion Transfer}

Video generation using conditional GANs is also done using additional motion or pose information from image sequences \cite{hu2018video,ohnishi2017hierarchical,chan2019everybody}. %\cite{hu2018video} uses first input frame and a pre-computed motion stroke denoting the approximate motion of the object to synthesize a video. 
\cite{ohnishi2017hierarchical} uses optical flow and synthesizes realistic images. \cite{chan2019everybody,yang2018pose} use pose information for motion transfer between videos. Extracting this motion can be a limitation for these approaches. Our method instead uses a generative approach where the motion can be synthesized instead of using a driving video.

\iffalse
Our work is related to these works in video synthesis as we are also using appearance as a prior while synthesizing the video. However, our problem is different from these in three key aspects. Firstly, these methods perform image synthesis and transfer the motion from source to target image one frame at time whereas we synthesize the full action video at once which is much more efficient. Secondly, whereas most methods do not make use of action semantics which can be used for generating good action features, we make use of this semantic information as an action prior. Lastly, we use a novel dual discriminator adversarial approach wherein we use a novel mix-adversarial loss which is specifically designed for video generation.
\fi

\section{Approach}

\begin{figure*}
\begin{center}
% \fbox{\rule{0pt}{2in} \rule{.9\linewidth}{0pt}}
\includegraphics[width=0.85\linewidth]{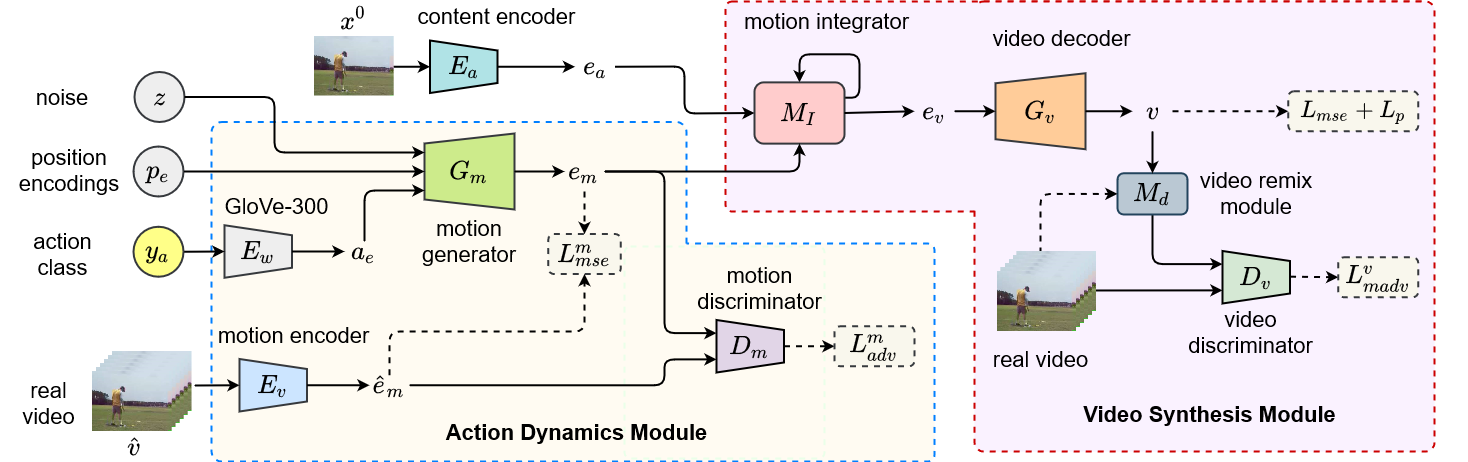}
\end{center}
  \caption{%Overview of the video generator1: Green: GRU, Blue:Upsampling conv block, Yellow: Normal conv block, arrows indicate concatenation of input to GRU with output, last upsampling layer is only in the spatial dimension
  \small{Overview of the proposed framework. Given an actor image $x^0$, action class $y_a$, position encoding $p_e$, and noise $z$, the network generates corresponding action video $v$. The motion generator $G_m$ generates action representation $e_m$ in latent space in the action dynamics module. Next, the motion integrator $M_I$ recurrently integrates $e_m$ with the appearance $e_a$ in a latent space to produce video features $e_v$ which is used to synthesize the action video $v$. The complete network is trained end-to-end with the help of multiple objective functions.
  }}
\label{fig:framework}
\end{figure*}

Given an input image $x^0$ of an actor and an action class $y_a$, our goal is to predict a video $v$ with $T$ frames $x^1, x^2, ..., x^T$ depicting how the action will be performed.
%Given an input image of an actor and an action class, our goal is to predict a sequence of video frames depicting how the action will be performed. Formally, given an input image of an actor $x^0$ and any action class $y_a$, we aim to generate a video $v$ with $T$ frames $x^1, x^2, ..., x^T$. 
To solve this, we propose LARNet consisting of two main parts; 1) \textit{action dynamics module}, and 2) \textit{video synthesis module} (Figure \ref{fig:framework}). The \textit{action dynamics module} generates latent action representation $e_m$ and the \textit{video synthesis module} synthesizes the action video $v$ by integrating the appearance and generated motion features. % An overview of the proposed framework is shown in Figure \ref{fig:framework} and we describe its components in the following subsections.

% There are several challenges involved in this problem. It is not trivial to estimate the dynamics of any action from its name and there can be a lot of variations in how a given action can be performed. Also, synthesizing a temporally coherent video can be challenging due to its spatio-temporal complexity and high memory requirement. 

% \vspace{-0.2cm}

\subsection{Action Dynamics Module}
\label{sec:act_rep}

It has been shown that decoupling appearance and motion component of a video provides more flexibility and improves overall quality for video synthesis \cite{Tulyakov_2018_CVPR_mocogan,wang2020g3an,huang2018multimodal}. Motivated by this, we propose to model the action dynamics as a latent representation conditioned on the action, independent of the appearance, by using a \textit{motion generator} $G_m$, which utilizes the action class label $y_a$ to estimate the latent action representation $e_m$. % To account for the variations in the performed action, we incorporate a stochastic latent noise $z$ that implicitly captures this variation.

Generating action dynamics merely based on a class label can be challenging. Therefore we develop a generative approach where we propose to approximate the generated action representation $e_m$ to motion features $\hat{e}_m$ extracted from a real action video $\hat{v}$. We use a 3D convolution based network $E_v$ to extract motion features from a video \cite{carreira2017quo}. To account for the temporal as well as action dynamics variations, we provide a position encoding $p_e$ and stochastic noise $z \sim \mathcal{N}(0, 1)$  to $G_m$. For position encoding $p_e$, we use the relative position of the starting frame of the action video $\hat{v}$ which is computed as a ratio of the frame position to the total number of frames in the video. The position encoding makes the learned action representation aware of the temporal variation in the action.

We use action semantics instead of a 1-hot encoding where the action name is converted to word embeddings $a_e = E_w$ with the help of GloVe-300 representation \cite{pennington2014glove}. 
%If an action name has multiple words, we add the embeddings, which is an effective way to extract semantics \cite{pennington2014glove}. 
The semantic encodings perform slightly better than 1-hot and enable the model to also synthesize novel actions. The motion generator $G_m$ is a 3D convolution based network which takes the semantic embeddings $a_e$, position encoding $p_e$ and stochastic noise $z$ as input and generates latent action representation $
    e_m = G_m(a_e, p_e, z)
$.

\subsection{Video Synthesis Module}

The \textit{video synthesis module} consists of two components; 1) \textit{motion integrator}, and 2) \textit{video decoder}. 
%To generate a temporally coherent video, we propose a hierarchical recurrent \textit{motion integrator} $M_{i}$, which integrates the motion representation $e_m$ with the actor appearance $e_a$ in latent space. The \textit{motion integrator} $M_{i}$ attempts to generate latent video representations in a recurrent process. The hierarchical structure of $M_{i}$ helps in generating video features at varying scales capturing both coarse level as well as fine level features. The video features $e_v$, incorporating both appearance $e_a$ and motion $e_m$, are used to generate the video $v$ with the help of a \textit{video decoder} $G_v$. 
The generated action representation $e_m$ is integrated with the appearance prior $e_a$ in a latent space to produce video features $e_v$ using a motion integrator $M_I$. We propose a recurrent approach which utilizes the generated action representation $e_m$ and transforms the appearance $e_a$ one step at a time according to the learned action. The motion integration module $M_I$ has a recurrent structure based on convolutional Gated Recurrent Unit (Conv-GRU) \cite{ballas2015delving}, which takes the encoded prior $e_a$ as input along with the generated action representation $e_m$ and predicts integrated video features $e_v$. Formally, $
    e_v = M_I(E_a(x^0), G_m(a_e, p_e, z))
$
where, $x^0$ is the actor image and $E_a$ is the image encoder where we use a 2D conv network \cite{simonyan2014very}.

The motion integrator takes the appearance latent representation $e^{t-1}_a$ at each time step and transforms it to $e^{t}_a$ using the latent action representation. First the foreground $f^t$ and background $b^t$ is separated based on the appearance $e^t_a$ and action features $e^t_m$ using learnable 2D kernels $W_f$ and $W_b$ respectively, $b^t = \sigma (W_{b}*<{e}^{t-1}_a, e_m^t>$. Then the background features $b^t_f$ are extracted based on prior latent appearance $e^{t-1}_a$. The foreground features $f^t_f$ are transformed using action kernels $W_a$ and action features $e^t_m$. Both foreground and background features are combined to get the generated video features $e^t_v$ for time step $t$, given as
\begin{equation}
\small
{e}^t_v = (b^t \odot e_a^{t-1}) +  
    (1-b^t) \odot [tanh(W_a*<e_m^t, (\sigma (W_{f}*<{e}^{t-1}_a, e_m^t>)) \odot e_a^{t-1}>)].
\end{equation}
Lastly, the generated video features at each time step are combined together to form integrated video features $e_v$ and video is generated via a video decoder $G_v$.
% \subsection{Video Decoder}
The integrated latent video features $e_v$ are used to generate a video $v$ where the actor present in the image prior $x^0$ performs the target action $y_a$. Formally this can be described as,
\begin{equation}
\small
    v = G_v(M_I(E_a(x^0), G_m(a_e, p_e, z))).
\label{integrator}
\end{equation}
where the generated latent action representation $e_m = G_m(a_e, p_e, z)$  is integrated with the latent appearance $e_a = E_a(x^0)$ to generate the required video $v$ with the help of a video decoder $G_v$ which is a 3D convolution based network. % It comprise of convolution layers in combination with upsampling layers where we use trilinear interpolation.

\paragraph{Hierarchical Motion Integration}

%While the motion integrator $M_I$ generates the latent video features $e_v$ by transforming the latent action representation, the encoding compression obscures fine action details. To overcome this challenge, 
To improve on fine action details lost during action encoding, we propose to integrate the motion with appearance at multiple scales using a hierarchical motion integrator, generating coarse to fine features accordingly. In each level, the motion integrator $M_I$ takes the latent appearance features $e_a$ and action representation $e_m$ along with generated video features from previous level and generates video features $e_v$ which are then passed to a video decoder to generate higher resolution features. 
%The generated features are combined with the integrated video features from the next level motion integrator $M_I$ which are again passed to the video decoder. The higher level appearance features are extracted from the earlier layers of the appearance encoder $E_a$. 
Similarly, the action generator $G_m$ is trained to generate action features at multiple resolutions.
%and corresponding motion features are extracted from earlier layers of motion encoder $E_v$. %More architectural details are provided in the supplementary.

\subsection{Training Objective}

% The motion generator $G_m$ aims at learning a latent action representation $e_m$, which should be a good approximation of motion features $\hat{e}_m$ extracted from a real action video. 
We use mean squared error loss $L^{m}_{mse}$ and adversarial loss $L^m_{adv}$ to learn a latent action representation. We use a 3D convolution based discriminator $D_m$ to differentiate between the generated representation $e_m$ = $G_m(a_e, p_e, z)$ and the motion representation $\hat{e}_m$ = $\mathcal{E}_v(\hat{v})$ extracted from a real video. 
%To further ensure that the motion generator $G_m$ generates realistic action dynamics in the latent representation, we use adversarial learning. A 3D convolution based discriminator $D_m$ is used where $D_m$ tries to differentiate between the generated representation $e_m$ = $G_m(a_e, p_e, z)$ and the motion representation $\hat{e}_m$ = $\mathcal{E}_v(\hat{v})$ extracted from a real video. 
The adversarial objective $L^m_{adv}$ is determined using a Wasserstein loss formulation with a gradient norm penalty \cite{gulrajani2017improved} for a stable network training. 
\begin{equation}
\small
\label{eq:wgan}
    L^m_{adv} = \mathop{\mathbb{E}}_{x \sim \mathbb{P}_g} [\mathcal{D}_m(x)] - \mathop{\mathbb{E}}_{\widetilde{x} \sim \mathbb{P}_r} [\mathcal{D}_m(\widetilde{x})] + \lambda \mathop{\mathbb{E}}_{\hat{x} \sim \mathbb{P}_{\hat{x}}} [(|| \nabla_{\hat{x}} \mathcal{D}_m(\hat{x})||_2 - 1)^2]. 
\end{equation}
Here $\mathbb{P}_g$ represents generated representation, $\mathbb{P}_r$ represents extracted representations, and $\mathbb{P}_{\hat{x}}$ represents sampling along straight lines between pairs of points sampled from $\mathbb{P}_g$ and $\mathbb{P}_r$. $\lambda$ is the penalty coefficient which we set to 10 according to \cite{gulrajani2017improved}.

%Our overall training objective aims at generating realistic action video and we utilize multiple loss functions to achieve this goal. The first component is based on a reconstruction loss where we make use of mean squared error between the generated and the ground-truth video. The mean squared loss $L_{mse}$ encourages the generated video to match the ground-truth video $\hat{v}$. 
% \vspace{-0.3cm}

\paragraph{Mix-adversarial Loss}
%To help the generator predict realistic looking videos, we propose a novel formulation of adversarial loss which is specifically designed for videos. 
Differentiating between real and generated videos for a discriminator is often easier during initial stages and gradually gets harder as training progresses, which also causes generator saturation. We propose a remix strategy where the generated and real videos are fused together by a video remix module $M_d$ which stochastically remixes their frames. This mix-video $v_m$ is used as fake instead of the generated video for adversarial learning. The key idea is to introduce temporal inconsistency in the generated video, which improves discriminator performance and also forces the generator to synthesize a temporally coherent video. %The mix-video $v_m$ is created in a stochastic process where a time-step is randomly sampled and all frames before that time-step are replaced by the frames from ground-truth video. It is performed by a video remix module $M_d$ which takes both generated and real action video as input. 
The loss objective $L^v_{madv}$ for mix-adversarial learning is,
\begin{equation}
\small
\label{eq:wgan2}
    L^v_{madv} = \mathop{\mathbb{E}}_{x \sim \mathbb{P}_g, \widetilde{x} \sim \mathbb{P}_r} [\mathcal{D}_v(Mix(x, \widetilde{x}))] - \mathop{\mathbb{E}}_{\widetilde{x} \sim \mathbb{P}_r} [\mathcal{D}_v(\widetilde{x})] + \lambda {\mathop{\mathbb{E}}_{\hat{x} \sim \mathbb{P}_{\hat{x}}}} [(|| \nabla_{\hat{x}} \mathcal{D}_v(\hat{x})||_2 - 1)^2]
\end{equation}
Here $Mix()$ represents frame mixing of generated and real videos, $\mathbb{P}_g$ and $\mathbb{P}_r$ represents distribution of generated and real videos respectively, $\mathbb{P}_{\hat{x}}$ represents sampling along straight lines between pairs of points sampled from $\mathbb{P}_g$ and $\mathbb{P}_r$, and $\lambda$ is the penalty coefficient.% We use a 3D convolution based discriminator $D_{v}$ to distinguish between the fake mix-video and real action video. 

We use \textit{MSE} loss $L_{mse}$ between generated and real videos to push generator to create realistic videos and a perceptual loss $L_p$ to improve the its perceptual quality \cite{johnson2016perceptual}. The proposed framework is trained end-to-end and the overall training objective is,
\begin{equation}
\small
\label{eq_loss}
    L = \lambda_1L^{m}_{mse} + \lambda_2L^m_{adv} + \lambda_3L_{mse} + \lambda_4L^v_{madv} + \lambda_5L_{p},
\end{equation}
where $\lambda_1, \lambda_2,\lambda_3, \lambda_4$, and $\lambda_5$ are weights which are determined experimentally.

\section{Experiments}

% \subsection{Experiment Setup}

%The proposed approach jointly optimizes multiple objectives and is trained end-to-end using a combined loss $L$. 
We demonstrate the effectiveness of the proposed approach and highlight the benefits of its main components (action representation learning, hierarchical motion integrator, and mix-adversarial loss) via quantitative and qualitative evaluation. 
% We use 0.25 and 0.3 for perceptual loss ($\lambda_5$) and representation mse loss ($\lambda_1$) respectively with all other loss terms with weight 1 in equation \ref{eq_loss} to estimate $L$. We employ Adam optimizer \cite{kingma2014adam} with a learning rate of 1e-4 and $\beta_1 = 0.5$ and $\beta_2 = 0.9$ and train using the PyTorch framework on a Nvidia Titan-RTX 24GB with a batchsize of 12. More details are in supplementary. 

% \paragraph{Datasets}

%We experimented with five real-world human action datasets including NTU-RGB+D \cite{shahroudy2016ntu}, UCF-101 \cite{soomro2012ucf101}, Penn Action \cite{zhang2013actemes} , KTH \cite{schuldt2004recognizing} and UTD-MHAD \cite{chen2015utd} and a synthetic dataset.
We experimented with four real-world human action datasets including NTU-RGB+D \cite{shahroudy2016ntu}, Penn Action \cite{zhang2013actemes} , KTH \cite{schuldt2004recognizing} and UTD-MHAD \cite{chen2015utd} with a resolution of 112x112.  

\paragraph{Evaluation Metrics}

We evaluate the quality of the generated videos using frame level Structural Similarity Index Measure (SSIM) \cite{wang2004image} and Peak Signal to Noise Ratio (PSNR) \cite{hore2010image} against the ground-truth video. Apart from these, we also evaluate the realism of the generated videos using video level FVD \cite{unterthiner2018towards,unterthiner2019fvd}, frame level FID \cite{heusel2017gans} scores. %  , and Inception scores \cite{salimans2016improved}. 

% \YSR{are we also evaluating on IS and LPIPS? If yes, cite them here...}

\paragraph{Baselines}

%To highlight the importance to LARNet we compare with some baselines. 
Our first baseline, BaseNet-1, does not use action dynamics module and the proposed motion integrator. It directly uses the action class instead and performs a joint content and motion learning (Figure \ref{fig:teaser} (a)). A second baseline, BaseNet-2, utilizes the action dynamics module without any explicit supervision (Figure \ref{fig:teaser} (b)). Our third baseline, LARNet-Base, uses a supervision on generated action representation (Figure \ref{fig:teaser} (d)). % It is trained using a mean squared error objective for both action representation and video synthesis.

\begin{table}[t!]
\begin{center}
\footnotesize
\begin{tabular}{c c c c c c c }
%\hline
\textbf{Method} & \textbf{Driving Video} & \textbf{PSNR} $\uparrow$ & \textbf{SSIM} $\uparrow$ & \textbf{FID} $\downarrow$ & \textbf{FVD} $\downarrow$\\

\hline %\hline
VGAN \cite{vondrick2016generating} &  & 15.8 & 0.74 & 181.29 & 15.36 \\
%\hline
MoCoGAN \cite{Tulyakov_2018_CVPR_mocogan} &  & - & - & 229.26 & 16.37\\
%\hline
G3AN \cite{wang2020g3an} &  & - & - & 183.08 & 17.13 \\
%\hline
Monkey-Net \cite{siarohin2019animating_monkeynet} & \checkmark & - & - & 215.23 & 22.61 \\
%\hline
Imaginator \cite{WANG_2020_WACV} &  & 26.1 & 0.93 &  157.31 & 15.90 \\
\hline
\textbf{LARNet\textsuperscript{\rm \dag} (Ours)} & \checkmark & 28.1 & 0.94 & 166.34 & 13.45 \\
\textbf{LARNet\textsuperscript{\rm \dag \dag} (Ours)} & \checkmark & 28.4 & 0.94 & 171.34 & 13.60 \\ 
\textbf{LARNet (Ours)} &  & \textbf{28.4} & \textbf{0.94} & \textbf{164.53} & \textbf{12.91} \\

\end{tabular}
\end{center}
\caption{\small{Comparison with existing conditional video synthesis methods on the NTU-RGB+D dataset. \textbf{\rm \dag} and \textbf{\rm \dag \dag} uses motion from a driving video where \textbf{\rm \dag \dag} uses a driving video instead of generated action during inference while \textbf{\rm \dag} is trained using a driving video without action dynamics module.
% uses driving video directly at test time while \textbf{\rm \dag \dag} uses driving video .
% \YSR{you don't have to say NTU, all scores are on NTU. remove that column... maybe add one column which says driving video... why don't you have psnr/ssim for monkey-net?}
% \AR{Monkeynet uses driving video and morphs the source image onto it, so there is no single video to compare psnr/ssim. We use only embedding from driving video, but they morph each frame of driving video directly. We could remove psnr/ssim from our as well}
}
}
\label{table_quantitative_NTU}
\end{table}

\begin{table}[ht!]
\begin{center}
\footnotesize
\begin{tabular}{c c c c c c }
%\hline
\textbf{Method}& \textbf{Dataset}  & \textbf{PSNR} $\uparrow$ & \textbf{SSIM} $\uparrow$ & \textbf{FID} $\downarrow$ & \textbf{FVD} $\downarrow$\\
\hline
G3AN \cite{wang2020g3an} & Penn  & - & - & 63.1 & 24.24 \\
Imaginator \cite{WANG_2020_WACV}& Penn &19.3  &0.69& 64.8 & 13.88   \\
%\hline
\textbf{LARNet (Ours)}  & Penn  &\textbf{23.6 } &\textbf{0.80}& \textbf{52.1} & \textbf{8.45}  \\
\hline
%\hline

G3AN \cite{wang2020g3an} & UTD  & - & - & 87.6 & 17.8\\
Imaginator \cite{WANG_2020_WACV}& UTD   &{ 28.3} & {0.93} & {92.3}  & {19.3}\\ 
%\hline
\textbf{LARNet (Ours)}  & UTD  &\textbf{29.7 } &\textbf{0.94} & \textbf{ 77.1 }&\textbf{ 16.2}  \\ 
\hline 
%\hline

G3AN \cite{wang2020g3an} & KTH  & - & - & 173.7 & 24.1 \\
Imaginator \cite{WANG_2020_WACV}& KTH  &25.1  & 0.82& 127.6 & 15.5  \\
%\hline
\textbf{LARNet (Ours)}  & KTH &\textbf{26.6 }& \textbf{0.87} &  \textbf{104.3} & \textbf{15.3}  \\

\end{tabular}
\end{center}
\caption{\small{Comparison with existing conditional video synthesis methods. % on Penn Action, UTD-MHAD and KTH datasets.
}
% \AR{FID/FVD for G3AN and our method has to follow same protocol. \YSR{you can put the dataset name on top as an additional row, it will add three more rows, but will be fine.. also how about G3AN on KTH?}}. %\YSR{I think this could be one of the main reason for bad reviews, this table shows only our results...}
}
\label{table_quantitative_others}
\end{table}

% ==================================
% \begin{table}[t!]
% \begin{center}
% \footnotesize
% \begin{tabularx}{\linewidth}{l  @{\extracolsep{\fill}} c c c c c}
% %\hline 
% %& \textbf{NTU} & \textbf{NTU} & \textbf{NTU} \\
% \textbf{Method} & \textbf{PSNR} $\uparrow$ & \textbf{SSIM} $\uparrow$  & \textbf{FID} $\downarrow$ & \textbf{FVD} $\downarrow$ \\
% \hline %\hline
% VGAN \cite{vondrick2016generating}  & 15.8 & 0.74 & 181.29 & 15.36 \\
% %\hline
% MoCoGAN \cite{Tulyakov_2018_CVPR_mocogan}  & - & - & 229.26 & 16.37\\
% %\hline
% G3AN \cite{wang2020g3an} & - & - & 183.08 & 17.13 \\
% %\hline
% Imaginator \cite{WANG_2020_WACV}   & 26.1 & 0.93 &  157.31 & 15.90 \\
% \hline
% LARNet (Ours) & \textbf{28.4} & \textbf{0.94} & { 164.53} & \textbf{12.91} \\
% %\hline
% \end{tabularx}
% \end{center}
% \caption{Comparison with existing conditional video synthesis methods on NTU-RGB+D dataset. 
% %\YSR{Imaginator is doing better in terms of FID? fix the bold... we may want to explain that this in the text why they have better FID which is frame specific...}
% }
% \label{table_quantitative_NTU}
% \end{table}

% ==================================

\begin{figure}[t!]
\begin{center}
% \fbox{\rule{0pt}{2in} \rule{.9\linewidth}{0pt}}
\includegraphics[width=0.85\linewidth]{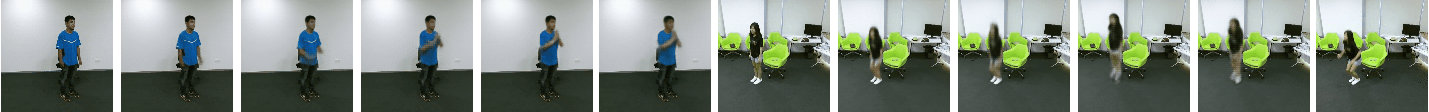}
\includegraphics[width=0.85\linewidth]{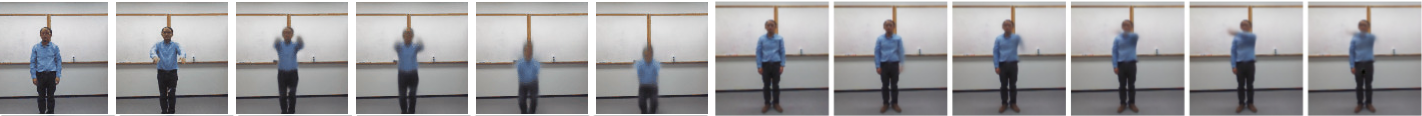}
\includegraphics[width=0.85\linewidth]{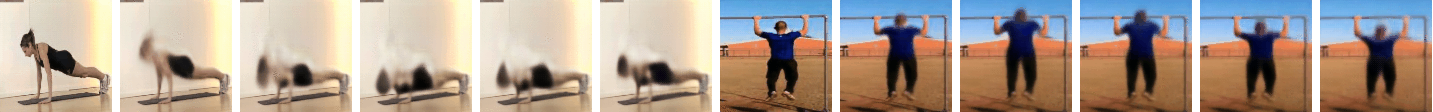}
% \vspace{0.2cm}
%\includegraphics[width=1\linewidth]{latex/images/results/ucf2.png}
% \includegraphics[width=0.32\linewidth]{latex/images/results/ucf.png}
\includegraphics[width=0.85\linewidth]{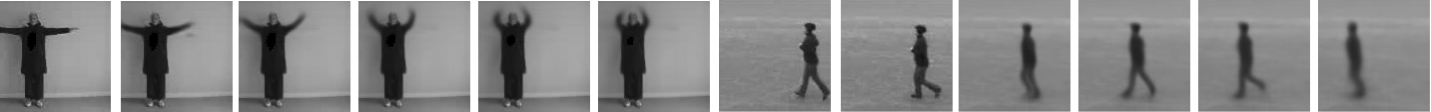}
\end{center}
  \caption{\small{Generated action videos on four different datasets using LARNet including NTU-RGB+D (row 1), UTD (row 2), Penn Action (row 3), and KTH (row 4).}}
\label{fig:gan_gen_all_datasets}
\end{figure}

% ==================================

% ==================================

\begin{figure*}[t!]
\begin{center}
% \fbox{\rule{0pt}{2in} \rule{.9\linewidth}{0pt}}
\includegraphics[width=0.85\linewidth]{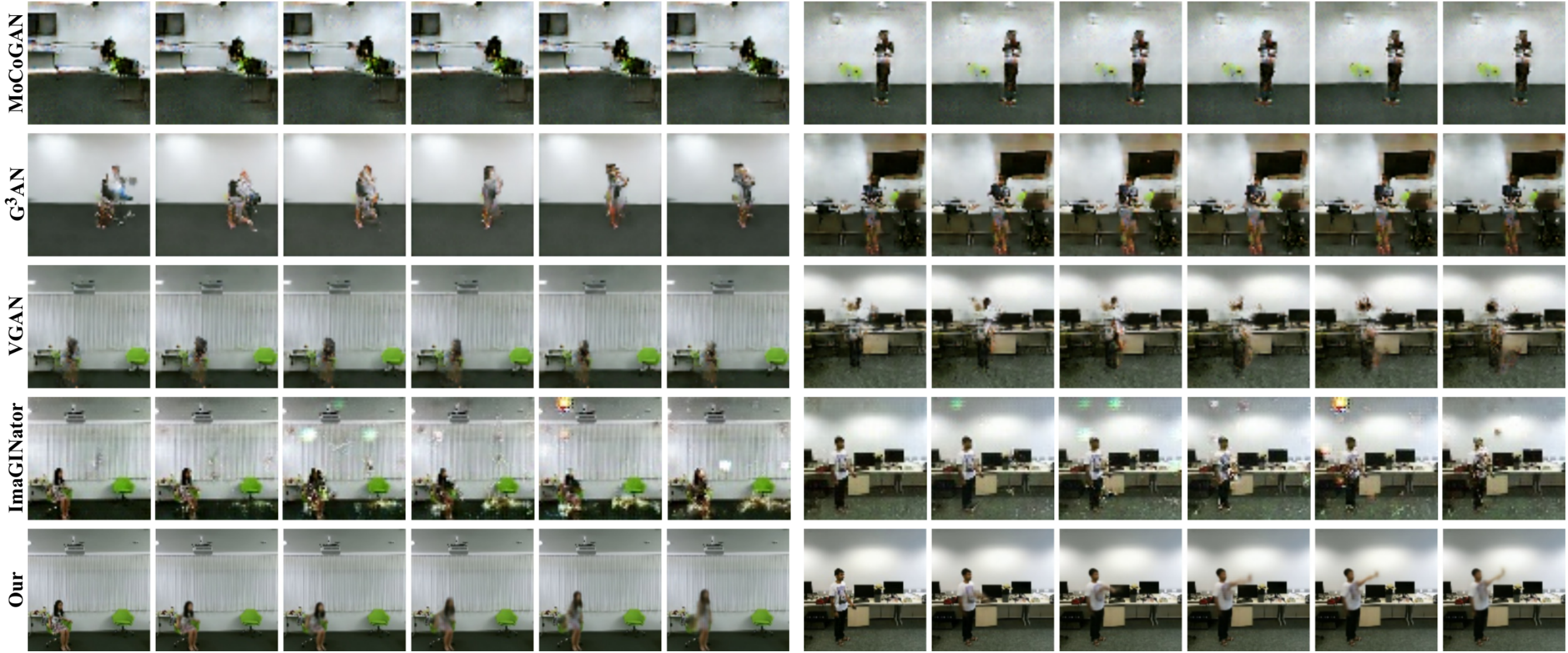}
\end{center}
  \caption{\small{Generated results compared with VGAN \cite{vondrick2016generating}, MoCoGAN \cite{Tulyakov_2018_CVPR_mocogan} , G3AN \cite{wang2020g3an}, and Imaginator \cite{WANG_2020_WACV} on NTU-RGB+D dataset.}
%   \YSR{it seems this and next figures are same thing... probably combine these two figures. no need to show real here... the first three samples are good, try to get another one for fourth, its not good for us... you can rotate labels vertically and show 2 rows for each method, 4 samples total. no need to put that dashed line, just increase the spacing, also increase the spacing between rows of different method... try to put 8 frames instead of 6... and also try to find visually similar samples from other methods (maybe later when everything else is done)}
  }
\label{fig:ntu_generated_prior}
\end{figure*}

\subsection{Evaluation on Human Actions}

We further evaluate LARNet on four different real-world human action datasets. The computed PSNR, SSIM, FID, and FVD scores are shown in Table \ref{table_quantitative_NTU} and \ref{table_quantitative_others}. The generated videos on four different action datasets using LARNet are shown in Figure \ref{fig:gan_gen_all_datasets}. We observe that the generated videos capture the action dynamics for a wide range of human actions. This is true even for those actions where only a slight movement of arms is involved,  such as ‘hand waving’ and ‘eating’. We also observe that the quality of the generated action videos is much better for UTD when compared with other datasets such as NTU-RGB+D. This can be explained by the complex scene structure and lot of action variations in NTU-RGB+D dataset. Next, we compare the quantitative and qualitative performance of LARNet with existing conditional video synthesis methods, including VGAN \cite{vondrick2016generating}, MoCoGAN \cite{Tulyakov_2018_CVPR_mocogan}, G3AN \cite{wang2020g3an}, and Imaginator \cite{WANG_2020_WACV}.

\paragraph{Quantitative Comparison}

We first compare the performance on NTU-RGB+D dataset, which is one of the largest human action dataset, in Table \ref{table_quantitative_NTU}. Our method outperforms all the other approaches in terms of PSNR, SSIM, and FVD scores. We observe that Imaginator \cite{WANG_2020_WACV} has a slightly better performance in terms of FID score which could be due to the use of frame level adversarial loss. However, it is important to note that FID only measures frame level quality whereas FVD is more focused on video dynamics. LARNet outperforms other methods in terms of FVD score. 

% Next we compare the performance on UCF-101 which is more challenging due to complex scenes. The comparison is shown in Table \ref{table_quantitative_UCF} and we observe that LARNet outperforms other methods in all the metrics. This illustrates that LARNet can generate better quality videos with varying action dynamics when compared with other methods. 

Next, we compare the performance on small scale datasets including Penn Action, UTD-MHAD and KTH to evaluate the generalization capability of LARNet. We compare with G3AN \cite{wang2020g3an} and Imaginator \cite{WANG_2020_WACV} in Table \ref{table_quantitative_others}. Even on small sized datasets LARNet consistently outperforms these two methods on all four metrics. 

% The method provides a FID score of 165.52, a FVD score of 13.34, and an inception score of 8.52. The PSNR and SSIM scores are 28.4 and 0.94 respectively which demonstrates the effectiveness of the proposed method in action prediction. The quantitative evaluation on UCF-101 dataset is shown in Table \ref{table_quantitative_ucf101}. UCF-101 dataset actions are very challenging as they are captured in the wild and have a lot of variation. Therefore, we observe that the performance on UCF-101 dataset is not as good as NTU-RGB+D dataset in terms of all the evaluation metrics. Next, we also evaluate the proposed method on smaller datasets, UTD-MHAD, and Penn Action. The results are shown in Table \ref{table_quantitative_ucf101} for all the metrics. We observe that results are better on UT-MHAD dataset when compared with Penn Action and this could be due to the complex scenes and actions present in Penn Action dataset.

\begin{figure}[t!]
\begin{center}
% \fbox{\rule{0pt}{2in} \rule{.9\linewidth}{0pt}}
\includegraphics[width=0.35\linewidth]{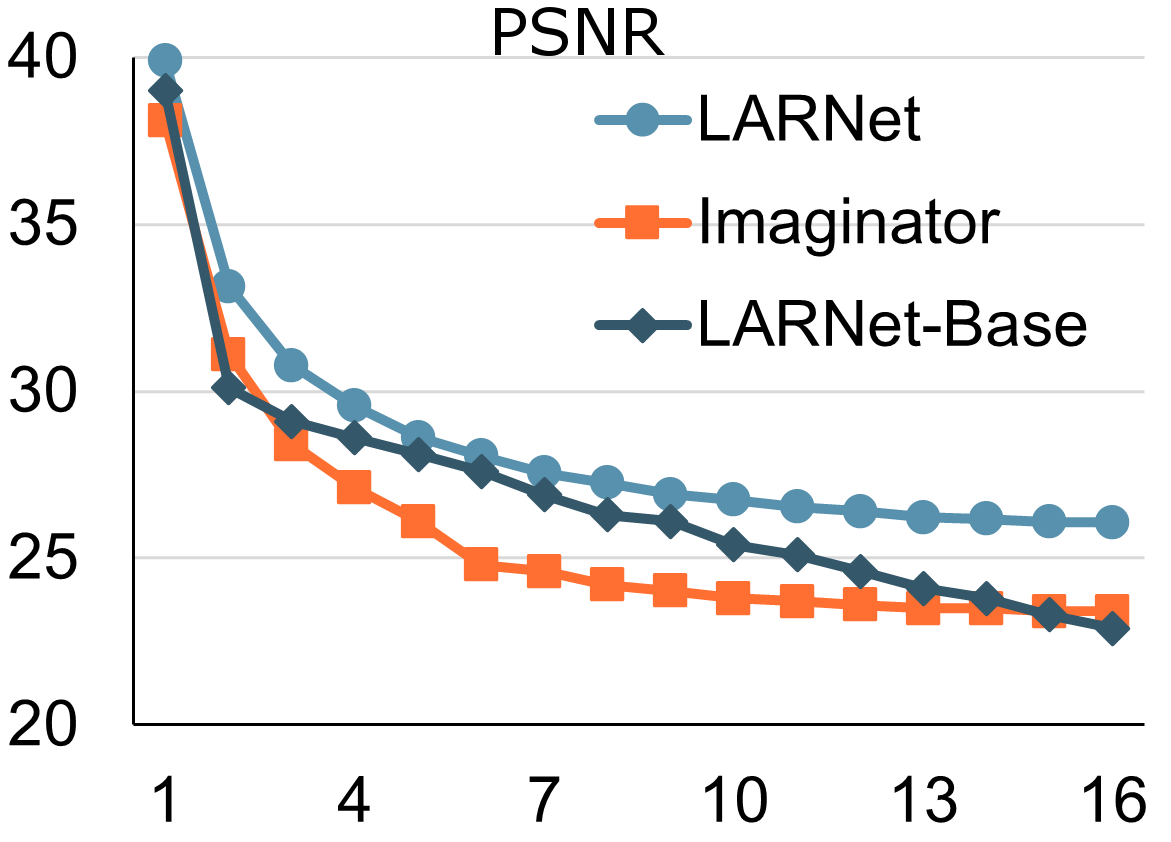}
\includegraphics[width=0.35\linewidth]{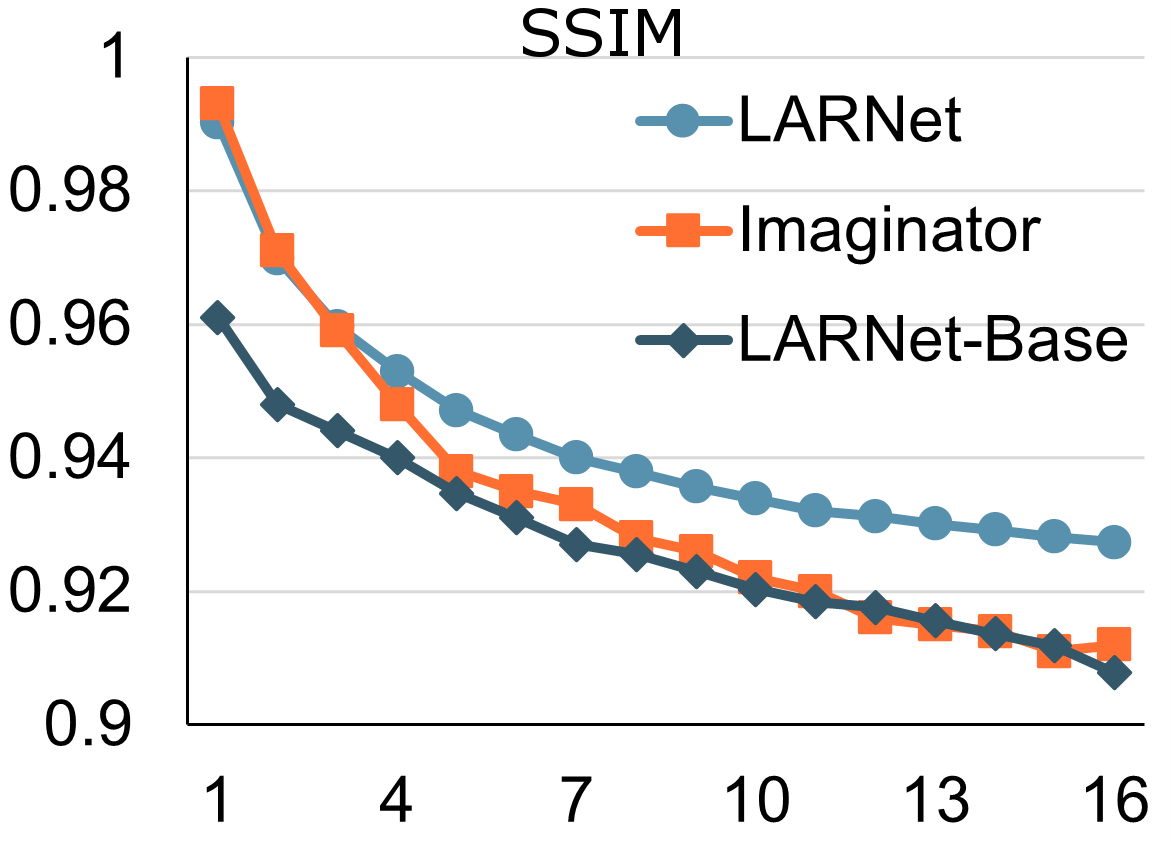}
\end{center}
  \caption{\small{Variation of frame quality in the generated videos with time. % The scores are averaged over the entire test set. % (Left: PSNR. Right: SSIM)
  }}
\label{fig:framewise_psnr}
\end{figure}

The proposed method generates 16 consecutive frames at a time. In the qualitative results, we observe that the visual quality of frames degrade over time as the action is being generated. To analyze this further, we compare the quality of generated frames independently at each time-step with Imaginator and our baseline model. We utilize PSNR and SSIM scores for this comparison and it is shown in Figure \ref{fig:framewise_psnr}. We observe that as we move temporally, the quality degrades for all models but with LARNet the quality is preserved much better than Imaginator which is mainly accredited to the hierarchical motion integrator which helps in preserving the fine level details.

\paragraph{Qualitative Comparison}

In Figure \ref{fig:ntu_generated_prior}, we show some generated videos for comparison with the existing methods. We observe that LARNet not only can keep better content information than other methods, but also captures the video dynamics for a wide range of actions. Although the other methods are able to generate a good background, they are not able to capture the fine level action details (such as motion of hands). These results show that LARNet can consistently generate the background content of still objects while synthesizing reasonable action dynamics, which clearly outperforms other methods.

\begin{table}[t!]
\begin{center}
\footnotesize
\begin{tabularx}{\linewidth}{l @{\extracolsep{\fill}} c c c c c}
%\hline
%& \textbf{NTU} & \textbf{NTU} & \textbf{NTU} \\
\textbf{Approach}  & \textbf{PSNR} $\uparrow$ & \textbf{SSIM} $\uparrow$& \textbf{FID} $\downarrow$ & \textbf{FVD} $\downarrow$  \\
\hline 
BaseNet-1  &26.1 & 0.919 & 67.1  & 14.18\\
%\hline
BaseNet-2 & 25.2  &0.912 &  65.3 & 14.21 \\
\hline
LARNet-Base & 26.41 & 0.921  &  66.5 & 14.15 \\
%\hline
LARNet-MI-1 & 26.83  & 0.927&  64.9& 14.11  \\
%\hline
LARNet-MI-3 & \textbf{27.25} &\textbf{ 0.931}  & \textbf{63.5} & \textbf{14.02}\\
\hline
\hline
LARNet-MI-3 + [$L^v_{adv}$]  & 27.23 & 0.933 & {63.1} & 13.89\\
% \hline
LARNet-MI-3 + [$L^v_{adv}, L^m_{adv}$]  &  {27.32} & 0.937 & 62.8 & 13.86\\
% \hline
LARNet-MI-3 + [$L^v_{madv}, L^m_{adv}$]  &  \textbf{27.39} & \textbf{{0.939}} & \textbf{62.2} & \textbf{13.71}\\
%\hline
% Full 112 res & 165.53 & 13.34 & & 28.4 & 0.94 \\
%\hline
\end{tabularx}
\end{center}
\caption{\small{Quantitative comparisons to study the effect of various components of LARNet and the effects of different loss terms on NTU-RGB+D dataset.
}}
\label{table_ablation_NTU}
\end{table}

\begin{table}[h!]
\begin{center}
\footnotesize
\begin{tabularx}{\linewidth}{l @{\extracolsep{\fill}} | c c c c}
\hline
\textbf{Approach} & \textbf{PSNR} $\uparrow$ & \textbf{SSIM} $\uparrow$ & \textbf{FID} $\downarrow$ & \textbf{FVD} $\downarrow$ \\
\hline \hline
$LARNet_{1-hot}$ & 27.11 & 0.93 & 168.34 & 14.09 \\
$LARNet_{GloVe}$ \cite{pennington2014glove} & \textbf{27.40} & \textbf{0.94} & 165.53 & \textbf{13.34}\\
$LARNet_{BERT}$ \cite{devlin2018bert} & 27.36 & \textbf{0.94} & \textbf{ 161.33} & 13.67 \\
\hline
\end{tabularx}
\end{center}
\caption{\small{Comparison of using different encodings for the action labels from NTU-RGB+D dataset in our LARNet model. $LARNet_{1 - hot}$ uses only one-hot encoding for the labels. $LARNet_{GloVe}$ uses the GloVe-300 text encoding for labels. $LARNet_{BERT}$ uses the BERT \cite{devlin2018bert} text encoding for labels. }}
\label{table_compare_onehot}
\end{table}

% \begin{table}[t!]
% \begin{center}
% \small
% \begin{tabularx}{\linewidth}{l @{\extracolsep{\fill}} c c c c c}
% %\hline
% %& \textbf{NTU} & \textbf{NTU} & \textbf{NTU} \\
% \textbf{Approach}  & \textbf{PSNR} $\uparrow$ & \textbf{SSIM} $\uparrow$& \textbf{FID} $\downarrow$ & \textbf{FVD} $\downarrow$  \\
% \hline %\hline
% LARNet-MI-3   & 27.25 & 0.931& 63.5 & 14.02\\
% \hline
% \hline
% +[$L^v_{adv}$]  & 27.23 & 0.933 & {63.1} & 13.89\\
% \hline
% % \hline
% +[$L^v_{adv}, L^m_{adv}$]  &  {27.32} & 0.937 & 62.8 & 13.86\\
% \hline
% +[$L^v_{madv}, L^m_{adv}$]  &  \textbf{27.39} & \textbf{{0.939}} & \textbf{62.2} & \textbf{13.71}\\
% % \hline
% % Patch &  27.38 & {0.939} & 62.5 & 13.73  \\
% % \hline
% % +Perceptual &  27.43 & {0.941} & 60.9 & 13.58  \\
% % \hline
% % $L^v_{adv}$ & {62.4} & {14.01} & {2.88} &  {27.35} & {0.938} \\
% % \hline
% %\hline
% % Full 112 res & 165.53 & 13.34 & & 28.4 & 0.94 \\
% \hline
% \end{tabularx}
% \end{center}
% \caption{Quantitative comparisons of our approach with different loss terms on NTU-RGB+D dataset.
% % \YSR{need to break this table into multiple to clearly show the effect of differnt components and loss terms, it is confusing now...}
% }
% \label{table_ablation_NTU_loss}
% \end{table}
% ==================================

% % ==================================

% \vspace{-0.3cm}

\subsection{Ablation Study}

We perform several ablation experiments to analyze the effectiveness of various components and loss functions in our approach. While the main experiments are done at 112x112 resolution, all the ablations are performed on NTU-RGB+D dataset at a resolution of 56x56. 

\paragraph{Effectiveness of Explicit Action Representation}
To evaluate the effectiveness of explicit action representation, we first train the proposed method without any motion generator (BaseNet-1). Next, we add the motion generator, but without any explicit supervision (BaseNet-2). Finally, we add a loss on the generated action representation for explicit modeling (LARNet-Base). The comparison is shown in Table \ref{table_ablation_NTU} and we can observe that adding explicit supervision outperforms both the variants in all metrics.

%We also compare the qualitative results for these ablations. The generated frames for these ablations are shown in Figure \ref{fig:ablation_ntu_generated}. We can observe that as we add these components to the BaseNet, the quality of generated frames improve. 

\paragraph{Effectiveness of Hierarchical Motion Integration}
To study the effect of hierarchical recurrent motion integrator $M_I$ on LARNet, we experimented with two different hierarchies on top of LARNet-base model. LARNet-MI-1 refers to recurrent motion integrator with single hierarchy and LARNet-MI-3 refers to a recurrent hierarchical motion integrator with three levels. A comparison of these two models is shown in Table \ref{table_ablation_NTU}. We observe that adding a three level motion integrator improves the PSNR and SSIM values as it focuses on both coarse level as well as fine level action dynamics.

% A comparative analysis of the performed ablations is shown in Table \ref{table_ablation_NTU}. We observe that adding each component improves the performance over the baseline network. However, the improvement is not consistent across different evaluation metrics. But it is important to note that different metrics measure different criterion's. The hierarchical motion integrator $M_I$ improves the PSNR and SSIM values as it focuses on both coarse level as well as fine level action dynamics. The adversarial loss terms on the other hand improves inception, FID and FVD scores, which are indicators of realism in the videos irrespective of what video is being generated. 

\begin{figure*}[h!]
\begin{center}
\includegraphics[width=.75\linewidth]{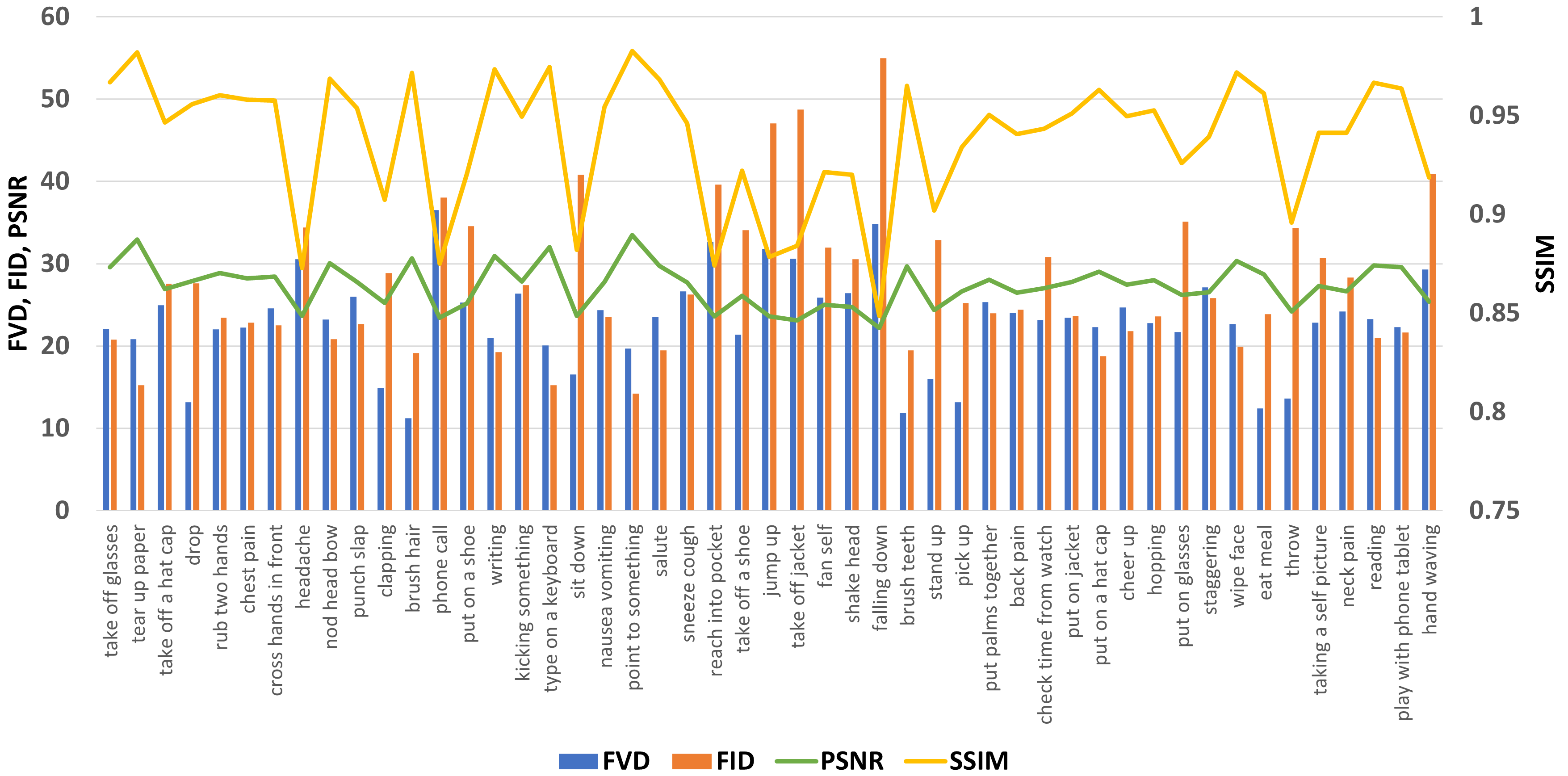}
\end{center}
  \caption{\small{Per class FVD, FID, PSNR and SSIM analysis on the NTU-RGB+D dataset. The left axis represents FVD, FID and PSNR scores while the right axis represents the SSIM scores per class.}
  }
\label{fig:ntu_per_class_scores}
\end{figure*}

\paragraph{Influence of Loss Functions}
To further demonstrate the effect of different loss functions we add a normal adversarial loss on LARNet-MI-3 model (+$L^v_{adv}$) for synthesized video. Next, we add an adversarial loss (+[$L^v_{adv}, L^m_{adv}$]) on generated motion dynamics. And finally, we use a mix-adversarial loss (+[$L^v_{madv}, L^m_{adv}$]) instead of normal adversarial loss on the generated video. A comparison of these loss functions is shown in Table \ref{table_ablation_NTU}. We observe that the adversarial loss terms improves FID and FVD scores, which are indicators of realism in the videos. Further, we also observe that the proposed mix-adversarial loss outperforms the classical adversarial loss in all four evaluation metrics. 

\paragraph{Effectiveness of different label encoding}
The effect of changing the label encoding on LARNet is shown in Table \ref{table_compare_onehot}, where we compare among using one-hot encoding, word encoding from GloVe-300 \cite{pennington2014glove} and word encoding from BERT \cite{devlin2018bert} models. We observe that the word encoding performs better than simple one-hot encoding in all metrics. While BERT encoding has better FID score, it performs slightly worse in FVD and PSNR metrics.

% ==================================

% \begin{figure}[t!]
% \begin{center}
% % \fbox{\rule{0pt}{2in} \rule{.9\linewidth}{0pt}}
% \includegraphics[width=0.75\linewidth]{latex/images/results/tsne.png}
% \end{center}
%   \caption{Tsne plot for model without contrastive(left) and with(right)
%   }
% \label{fig:tsne}
% \end{figure}
% \vspace{-0.27cm}

% \vspace{-0.4cm}

% \paragraph{Novel Actions}

% Same prior different action

% ==================================

% \begin{figure}[t!]
% \begin{center}
% % \fbox{\rule{0pt}{2in} \rule{.9\linewidth}{0pt}}
% \includegraphics[width=1\linewidth]{latex/images/results/syn3.png}
% \end{center}
%   \caption{Action semantics: top row - `left', second row - `bottom', and third row - `left bottom'.
% % \YSR{reduce the spacing for consistent space between images and its resolution, like previous figures...}
%   }
% \label{fig:syn3}
% \end{figure}

%\vspace{-0.5cm}

\subsection{Analysis and Discussion}

The proposed model generates video conditioned on the action type and to illustrate its effectiveness we use the same actor to synthesize different actions. In Figure \ref{fig:gan_gen_all_datasets} (row 2), we have shown two different generated videos using the same actor. We can observe that the action is distinctly visible in all the generated videos which demonstrates the capability of LARNet to synthesize diverse action videos.

\paragraph{Per Class Analysis}
We observe the per class quantitative performance on NTU-RGB+D dataset in Figure \ref{fig:ntu_per_class_scores}. It is observed that classes with lower FVD and FID scores (lower is better) also have higher PSNR and SSIM scores (higher is better), showing a correlation of improved performance across the metrics for those classes.

\paragraph{Limitations and Challenges}

%We have recently seen great progress in video prediction\cite{ye2019compositional,fragkiadaki2017motion,denton2018stochastic,walker2017pose, kumar2019videoflow, franceschi2020stochastic, xu2020video}, motion-transfer \cite{hu2018video,ohnishi2017hierarchical,chan2019everybody}, interpolation \cite{shen2020blurry,choi2020channel,bao2019depth}, and super-resolution \cite{haris2019recurrent,sajjadi2018frame,tao2017detail}. 
Video synthesis has been challenging as it needs an understanding of the action dynamics. Despite the recent efforts, the problem of video synthesis is still far from being solved. The proposed approach successfully generates videos with visible actions, however, modelling complex actions remains a challenge. Using significantly high computational resources have shown great improvement in this task by using large scale TPUs \cite{clark2019adversarial, villegas2019high}. Our training was limited to a single 24Gb GPU which we believe will scale well with the availability of higher computational resources.

\section{Conclusion}

In this work, we present a novel approach for generating human actions from an input image. The proposed framework predicts human actions conditioned on action semantics and utilizes a generative mechanism which estimates latent action representation. The latent action representation is explicitly learned with the help of a similarity and adversarial loss formulation. This learned latent representation is then used to generate an action video which is optimized using multiple objectives, including a novel mix-adversarial loss. We perform extensive experiments on multiple human action datasets demonstrating the effectiveness of various components of the proposed approach.

\bibliography{final_main_bmvc}

\end{document}